# Assessment of different loss functions for fitting equivalent circuit models to electrochemical impedance spectroscopy data

Ali Jaberi [3], Amin Sadeghi [2], Runze Zhang [1], Zhaoyang Zhao [1], Qiuyu Shi [1], Robert Black [3], Zoya Sadighi [3], Jason Hattrick-Simpers [1]

[1] Department of Material Science and Engineering, University of Toronto, Toronto, Ontario, Canada

[2] Canmet MATERIALS, Natural Resources Canada, 183 Longwood Road south, Hamilton, ON, Canada.

[3] Clean Energy Innovation Research Center, National Research Council Canada, Mississauga, Ontario, Canada

**Corresponding Author:** Jason Hattrick-Simpers

**Address:** Department of Material Science and Engineering, University of Toronto, Toronto, ON, Canada

**Email:** jason.hattrick.simpers@utoronto.ca

## Abstract

Electrochemical impedance spectroscopy (EIS) data is typically modeled using an equivalent circuit model (ECM), with parameters obtained by minimizing a loss function via nonlinear least squares fitting. This paper introduces two new loss functions, log-B and log-BW, derived from the Bode representation of EIS. Using a large dataset of generated EIS data, the performance of proposed loss functions was evaluated alongside existing ones in terms of $R^2$ scores, chi-squared, computational efficiency, and the mean absolute percentage error (MAPE) between the predicted component values and the original values. Statistical comparisons revealed that the choice of loss function impacts convergence, computational efficiency, quality of fit, and MAPE. Our analysis showed that X2 loss function (squared sum of residuals with proportional weighting) achieved the highest performance across multiple quality of fit metrics, making it the preferred choice when the quality of fit is the primary goal. On the other hand, log-B offered a slightly lower quality of fit while being approximately 1.4 times faster and producing lower MAPE for most circuit components, making log-B as a strong alternative. This is a critical factor for large-scale least squares fitting in data-driven applications, such as training machine learning models on extensive datasets or iterations.

# 1. Introduction

Electrochemical impedance spectroscopy (EIS) provides valuable insights into the kinetics and mechanisms of electrochemical systems and is extensively utilized in fields such as corrosion, semiconductor technology, energy conversion and storage, chemical sensing, and biosensing. The technique involves perturbing an electrochemical system at equilibrium or steady state by applying a sinusoidal signal (either alternating current or voltage) usually across a broad frequency between $10^{-3}$ and $10^6$ Hz and observing the system's corresponding response. The measurement usually records impedance data, which includes both real and imaginary parts that vary with frequency. [1,2]

EIS data are often interpreted using an equivalent circuit model (ECM), composed of circuit elements connected in various configurations that represent electrochemical processes. [1,2] The parameters of the circuit component are estimated by fitting the ECM to the EIS data. This process usually involves minimizing a loss function using nonlinear least squares (NLS) fitting. [2] Different loss functions have been used in the literature to find the optimum values of circuit parameters. The simplest loss function could be error vector without any weighting. However, when using unweighted loss functions, the parameters are estimated primarily by only the largest data values as the real and imaginary part of the impedance data often span over three or more orders of magnitude. [2–4] Thus, an appropriate weighting for loss function is usually necessary for estimating ECM parameters during NLS fitting.

Ideally, the loss function should be weighted based on its associated standard deviation either through inverse variance weighting [2,5] or measurements model [3,6–8]. However, this is not always practical, as it requires repeating impedance measurements multiple times. As a result, alternative weighting methods must be considered. Under the assumption that the errors in the impedance data are proportional to the magnitudes of the data values, a loss function with proportional weighting is usually suggested in the literature. [9,10] Zoltowski [11] and Boukamp [12,13] suggested that when the errors in real and imaginary residuals are not independent, a loss function weighted by modulus for both real and imaginary can be used. In this case, the loss function has the form of squared sum of weighted residuals known as chi squared.

Unfortunately, different loss functions can significantly impact the parameter values of the model.[9,10,14,15] For instance, Macdonald[14] demonstrated that when fitting impedance data of a hydrogen-doped lithium nitride with a two-time constant circuit, the non-Ohmic resistance value varied between 559 and 1122 ohms (i.e., approximately 50% difference) solely due to the choice of loss function. Using currently available loss functions for fitting EIS data, we also found that some are more effective at guiding the NLS fitting toward optimal component values before reaching the maximum number of allowed iterations. In addition, as data driven approaches become more popular, this lack of consistency of loss functions becomes problematic for training supervised classification or regression machine learning models for EIS applications. For example, using one loss function to generate labeled data and a different one during inference can significantly degrade model accuracy. This risk of accuracy degradation becomes even higher when the data generation and the model implementation are performed by different research groups using different loss function. In addition, selecting loss functions that reduce the computational cost of NLS fitting is crucial for improving training efficiency. Therefore, a key consideration is selecting a loss function that ensures consistency in circuit parameter estimates while balancing high-quality fits and minimizing computational cost.

The purpose of this paper is to evaluate the performance of various loss functions on synthetic EIS data. We proposed new loss functions based on the Bode representation of EIS which uses the relative magnitudes on the logarithmic scale and the phase differences. The performance of the existing and proposed loss functions was compared based on the $R^2$ scores, chi-squared and the computational efficiency. As part of our evaluation, we first investigated the loss functions on three case studies, followed by a statistical analysis of performance metrics for different loss functions. In addition, the mean absolute percentage error (MAPE) between the predicted component values and the original values used during data generation was computed for each loss function. We found that the choice of loss function plays a critical role in convergence, computational efficiency, and the quality of fit during the NLS fitting. Finally, based on these insights, we recommend suitable loss functions under different conditions, considering the performance criteria relevant to various research communities.

## 2. Theory and calculation

### 2.1. Loss Functions

The function that should be minimized in the least square procedure has a general form of: [2]

$$S = \sum_{i=1}^{n} \left\{ w_i^a \left[ f_{t,i}^a - f_m^a(\omega_i, \boldsymbol{P}) \right]^2 + w_i^b \left[ f_{t,i}^b - f_m^b(\omega_i, \boldsymbol{P}) \right]^2 \right\} \quad (1)$$

where $f$ is the impedance, the subscripts $t$ and $m$ stand for the true and model predicted data, a and b superscripts indicate two parts of the impedance data in either rectangular or polar domain (i.e. real/imaginary or magnitude/phase), $n$ is the number of data points, and $(\omega_i, \boldsymbol{P})$ emphasizes that the model values are a function of frequency and parameters. Commonly used loss functions can be derived from this equation by choosing $f$ and $w$. A complete list of loss functions investigated in this paper are presented in Table 1. If $f$ is represented in the rectangular form, then $f_i^a$ and $f_i^b$ correspond to the real ($Z_i'$) and imaginary ($Z_i''$) parts of the impedance. In this case, if $w = 1$, then the loss function simplifies to the squared sum of residuals with unity weighting (equation 1 in Table 1). When impedance magnitude is used as a uniform weight for both real and imaginary parts (i.e., $w_i^a = w_i^b = |Z_{t,i}|$), the loss function becomes the squared sum of residuals with modulus weighting (equation 2 in Table 1). If weights are proportional to the real and imaginary parts ($w_i^a = Z_{t,i}'$ and $w_i^b = Z_{t,i}''$), the loss function follows the squared sum of residuals with proportional weighting (equation 3 in Table 1). On the other hand, when $f$ represents impedance data in the polar form, then $f_{t,i}^a$ and $f_{t,i}^b$ correspond to the magnitude ($|Z_i|$) and phase ($\theta_i$). In this case, without additional weighting in the Bode representation, the loss function is the squared sum of Bode residuals (equation 4 of Table 1).

In addition to the aforementioned loss functions, two new loss functions are proposed here. These loss functions emphasize the Bode representation of EIS data and asses the residuals in the polar log-transformed domain. They measure the relative magnitudes on a logarithmic scale and the phase differences as the residuals between the true and predicted impedance data. As already mentioned, when the impedance scale varies substantially across frequencies, the weight becomes a vital part of the loss function. However, in this case the logarithmic transformation corrects the bias related to the scale during the least square procedure. The final loss function that should be minimized then becomes:

$$S_{\log-B} = \sum_{i=1}^{n} \left\{ \left[\log(|Z_{t,i}|) - \log(|Z_m(\omega_i, \boldsymbol{P})|)\right]^2 + \left[\theta_{t,i} - \theta_m(\omega_i, \boldsymbol{P})\right]^2 \right\} \quad (2)$$

where $|Z|$ is the magnitude of the impedance and $\theta$ is the phase. If the sole logarithmic transformation is not capable of correcting the bias related to the wide scale of data during the fitting process, the above equation can be further normalized by the logarithm of magnitude and phase as follows:

$$S_{\log-BW} = \sum_{i=1}^{n} \left\{ \left[\frac{\log(|Z_{t,i}|) - \log(|Z_m(\omega_i, \boldsymbol{P})|)}{\log(|Z_{t,i}|)}\right]^2 + \left[\frac{\theta_{t,i} - \theta_m(\omega_i, \boldsymbol{P})}{\theta_{t,i}}\right]^2 \right\} \quad (3)$$

**Table 1:** Loss functions used in this study

| Loss Function | Equation | Reference |
|---|---|---|
| 1. UW | $S_{UW} = \sum_{i=1}^{n} \left\{ [Z'_{t,i} - Z'_m(\omega_i, \boldsymbol{P})]^2 + [Z''_{t,i} - Z''_m(\omega_i, \boldsymbol{P})]^2 \right\}$ | 2,16 |
| 2. X2 | $S_{X2} = \sum_{i=1}^{n} \left\{ \left[\frac{Z'_{t,i} - Z'_m(\omega_i, \boldsymbol{P})}{|Z_{t,i}|}\right]^2 + \left[\frac{Z''_{t,i} - Z''_m(\omega_i, \boldsymbol{P})}{|Z_{t,i}|}\right]^2 \right\}$ | 11–13,17 |
| 3. PW | $S_{PW} = \sum_{i=1}^{n} \left\{ \left[\frac{Z'_{t,i} - Z'_m(\omega_i, \boldsymbol{P})}{Z'_{t,i}}\right]^2 + \left[\frac{Z''_{t,i} - Z''_m(\omega_i, \boldsymbol{P})}{Z''_{t,i}}\right]^2 \right\}$ | 9,10,17 |
| 4. B | $S_B = \sum_{i=1}^{n} \left\{ [|Z_{t,i}| - |Z_m(\omega_i, \boldsymbol{P})|]^2 + [\theta_{t,i} - \theta_m(\omega_i, \boldsymbol{P})]^2 \right\}$ | 2 |
| 5. log-B | $S_{\log-B} = \sum_{i=1}^{n} \left\{ [\log(|Z_{t,i}|) - \log(|Z_m(\omega_i, \boldsymbol{P})|)]^2 + [\theta_{t,i} - \theta_m(\omega_i, \boldsymbol{P})]^2 \right\}$ | Proposed loss function |
| 6. log-BW | $S_{\log-BW} = \sum_{i=1}^{n} \left\{ \left[\frac{\log(|Z_{t,i}|) - \log(|Z_m(\omega_i, \boldsymbol{P})|)}{\log(|Z_{t,i}|)}\right]^2 + \left[\frac{\theta_{t,i} - \theta_m(\omega_i, \boldsymbol{P})}{\theta_{t,i}}\right]^2 \right\}$ | Proposed loss function |

### 2.2. Fitting procedure

To determine the parameters of the ECM, either local or global optimization methods can be utilized. However, Z. Zhao et.al. [18] reported that global optimization approaches are generally unstable for this task and require substantially longer computational time than local methods, making them impractical for data-driven applications. In our evaluation, presented in the supplementary information, a global method consistently underperformed, further confirming these limitations. Therefore, this study employs a local optimization approach based on least-squares fitting, which is widely used within the community to determine the ECM parameters. [2] All the loss functions listed in Table 1 were tested in the least square procedure for 9000 synthetic data generated as described in section 2.3. The initial guesses for the parameters of the components are critical for convergence of least square, as they greatly impact the success of optimization algorithm for finding circuit parameters. [19,20] These initial guesses were randomly sampled within predefined value ranges. To have comparable computational efficiency, the random seeds are kept fixed for all fitting process. If the algorithm failed to meet the convergence threshold, new guesses were generated until either the convergence was achieved, or the number of iterations exceeded the maximum allowable value. The convergence threshold was set at a chi-squared value of 0.01, as an acceptable range for this metric [21], with an additional threshold of $R^2$ that was set to 0.9. While the loss functions in polar forms (B, log-B, and log-BW) minimize residuals in the Bode representation, chi-squared and $R^2$ were chosen to evaluate the fit quality in the Nyquist representation, as they are the standard metrics used in the field. The time for each fitting was also stored for computational efficiency comparison. **Algorithm 1** shows the fitting procedures used in this study. It is worth mentioning that the formulation of the chi-squared metric is the same as that in Equation 2 of Table 1. So, to avoid confusion throughout this paper, we use the symbol X2 when referring to the loss function, and we reserve the term 'chi-squared' for the performance metric.

---

**Algorithm 1** Pseudocode for least square procedure

---

1. **Initialize:**
   - Select loss function (obj)
   - Set thresholds for chi-squared and $R^2$ scores
   - parameters ← generate initial guess
   - iter ← 0
2. **while** true **do**
3.     iter ← iter + 1
4.     parameters ← least_squares(obj, parameters)
5.     chi-squared, $R^2$ ← calculate chi-squared and $R^2$ scores
6.     store the parameter set if the quality of fit is higher than all previous attempts
7.     **if** (chi-squared, $R^2$) meet the threshold or iter == max_iter **then**

|    8.     break
|    9.   end if
|    10.   parameters ← generate initial guess
| 11. end while
| 12. Compute final performance metrics, and time

## 2.3. Synthetic data

The ECMs used to generate synthetic EIS data are shown in Figure 1. In this figure, the *CPE*, *R*, *L* and *C* depict the constant phase element, the resistor, inductor and capacitor, respectively. For shorter notation, we will use *P* instead of *CPE* throughout this paper. The parameter ranges of the circuit components were selected to reflect battery and fuel cell systems. The ohmic resistance was sampled between 1 and 10 $\Omega$, while other resistances varied from 10 to $10^5$ $\Omega$. For *CPE*, the magnitude was sampled between $10^{-6}$ and $10^{-3}$ $\Omega$ $s^{-\alpha}$, with $\alpha$ (which quantifies deviation from ideal capacitance) sampled between 0.3 and 1. The value of *C* was also sampled between $10^{-6}$ and $10^{-3}$ $\Omega$ $s^{-1}$, while *L* was sampled between $1^{-6}$ and $1^{-3}$ $\Omega$ s. Finally, to simulate experimental noise, a Gaussian noise was infused to each component value at each impedance point. For each circuit, 1500 EIS data were generated, resulting in a total of 9000 EIS data.

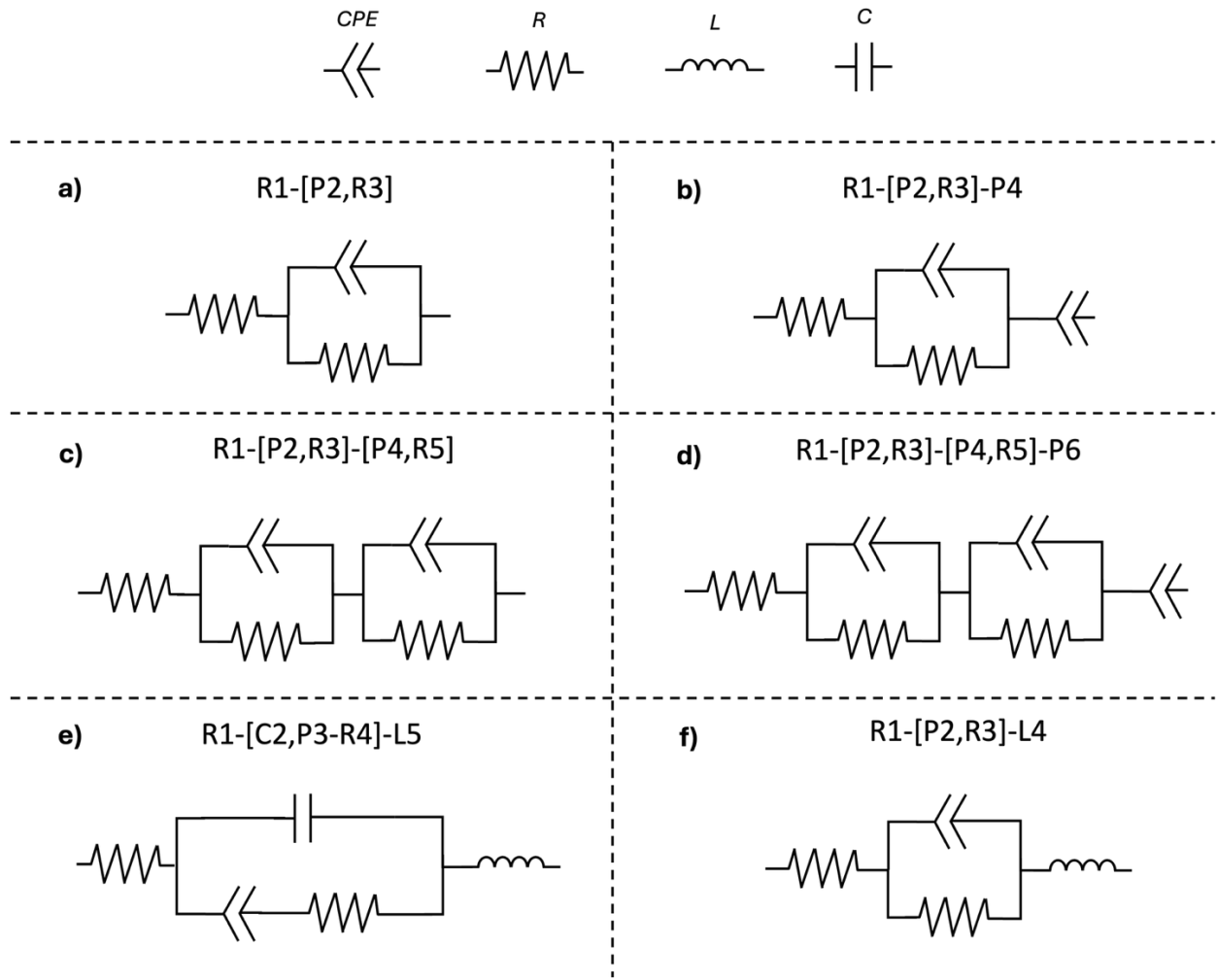

Figure 1: The six ECMs used for generating synthetic EIS data. The *CPE*, *R*, *L* and *C* depict the constant phase element, the resistor, inductor and capacitor, respectively. For shorter notation, *P is used instead of CPE* throughout this paper

## 3. Results and Discussion

Five performance metrics were considered to compare the loss functions. The computational time, chi-squared, three $R^2$ ($R^2$_score, $R^2$_magnitude, and $R^2$_phase) which were calculated for the complex rectangular domain, the magnitude and phase of the polar domain, respectively. Initially, three case studies were investigated followed by statistical comparison of these performance metrics and the MAPE between the predicted component values and the original values for large number of synthetic data.

### 3.1. Case studies

### 3.1.1. Case I: EIS data with narrow magnitude range

As the first case study, we chose an EIS from the data with narrow magnitude centered around $10^4$. This EIS which is shown in Figure 2, is a representative of the circuit illustrated in Figure 1-a which has values that produce narrow magnitude range. In Figure 2, the ECM-predicted impedance data after the least squares procedure using different loss functions are shown with orange triangles, and the actual representative EIS is shown with blue circles. If the least square procedure did not meet the convergence threshold before reaching the maximum allowable iterations, the ECM-predicted impedance corresponding to the iteration with the smallest error is shown in the figure. In this case, only the PW and log-BW loss functions could not meet the convergence threshold as is evident in the large deviations between ECM-predicted and actual EIS in Figure 2-b and Figure 2-c. Among the other loss functions that met the convergence, the X2 and log-B, as shown in Figure 2-c and Figure 2-e, respectively, demonstrated the highest quality of fit in terms of chi-squared. For a definitive conclusion, a more extensive statistical comparison is performed in in Section 3.2. We will show that the X2 and log-B outperform whereas loss functions with very large weighting, such as log-BW, overfit the impedance primarily in the high frequency range.

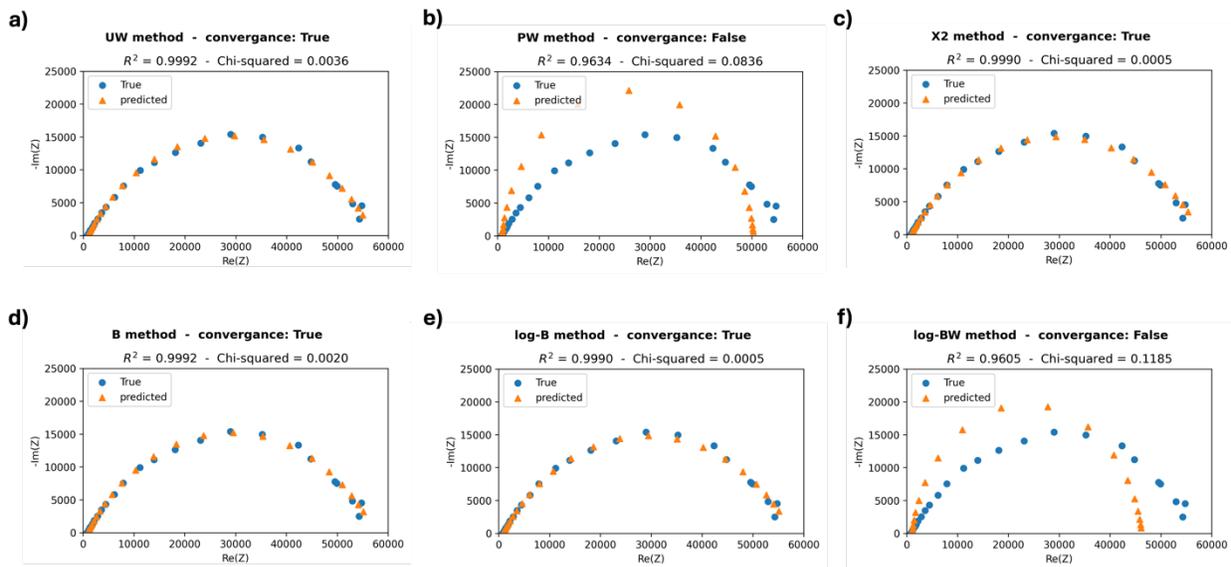

Figure 2: The true and the ECM-predicted impedance data fitted with different loss functions BW for circuit of Figure 1-a: a) UW, b) PW, c) X2, d) B, e) log-B, and f) log-BW

### 3.1.2. Case II: EIS data with wide magnitude range

As the second case study, we chose an EIS with wide magnitude range that is a representative of the circuit shown in Figure 1-b. The results of the least square fitting using different loss functions are illustrated in Figure 3 where the ECM-predicted impedance is shown in orange triangles, and the actual representative EIS is shown in blue circles. In this case, the UW and log-BW did not meet the convergence threshold. Interestingly as illustrated in Figure 3-a, the predicted EIS data using UW loss function still demonstrates a high-quality fit, achieving an $R^2$ score as high as 0.999, despite a poor chi-squared value of approximately 30. This highlights the limitation of relying on a single performance metric to assess fit quality for impedance data. A detailed examination of the zoomed-in high-frequency impedance data and the Bode representation further revealed significant discrepancies between the predicted and actual data, as shown in Figure 3-g and -h. Among the loss functions that met the convergence threshold, X2 and log-B outperformed the others in terms of the chi-squared and the $R^2$_score.

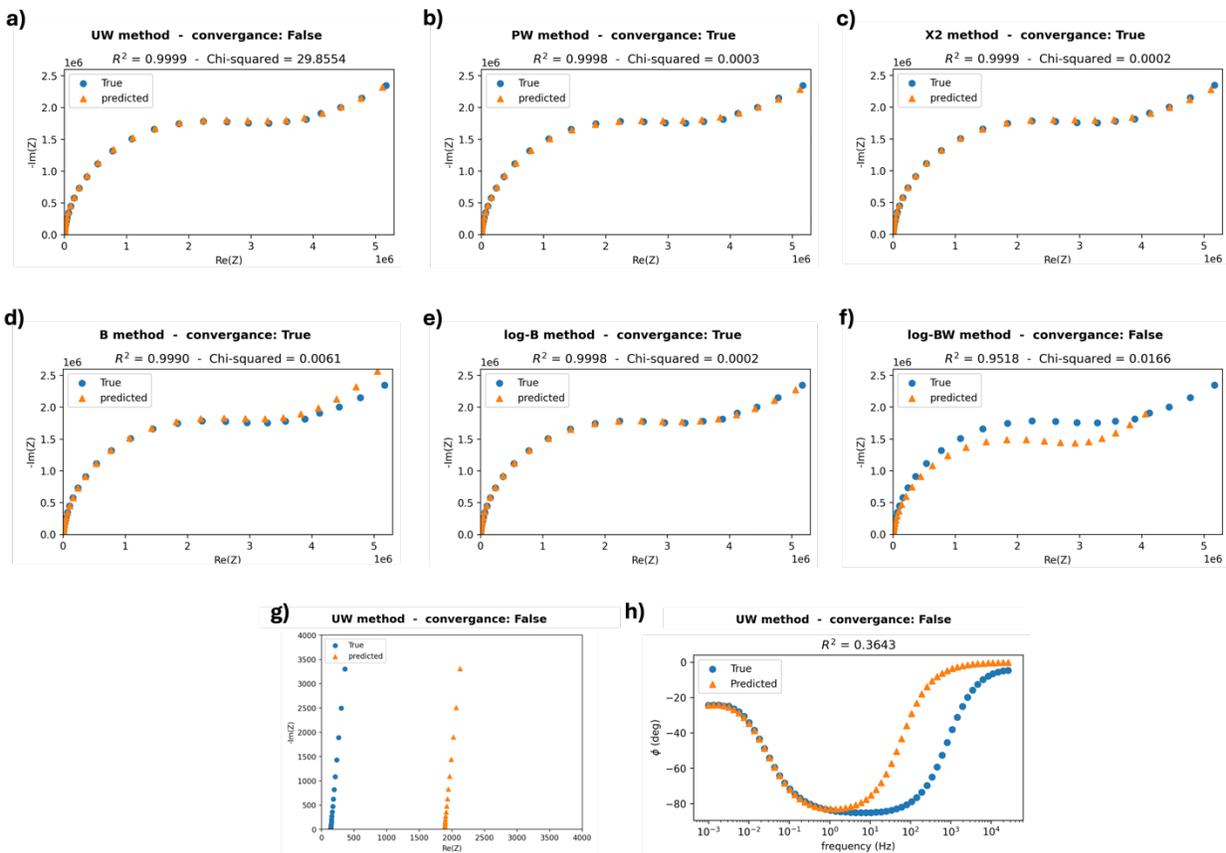

Figure 3: The true and ECM-predicted impedance data fitted with different loss functions for circuit of Figure 1-b: a) UW, b) PW, c) X2, d) B, e) log-B, and f) log-BW. Figure g) shows the zoomed-in Nyquist plot at high frequencies for UW loss function and figure h) shows the phase Bode plot for the UW loss function.

### 3.1.3. Case III: EIS data with wide magnitude range

As the third case study, we again chose an EIS with wide magnitude range that is a representative of the circuit shown in Figure 1-d. The results of the least square fitting using different loss functions are illustrated in Figure 4. In this case, only the log-BW loss function met the convergence threshold. This signifies that for some EIS data, a loss function with a very large weighting may be required to correct the bias related to the wide scale of data during the fitting process. It should be noted that while the magnitude range in case II is even wider than case III, the loss function with large weighting (i.e. log-BW) did not perform well in case II as observed in Figure 3-f. This could be possibly explained by distribution of magnitudes across the frequencies. Figure 5 shows the magnitude in Bode representation of EIS data and its distribution for both case II and case III fitted by the log-BW loss function. The magnitude for case II is mostly distributed around low value ranges and thus using a loss function with large weighting resulted in overfitting of data in high frequency region. On the other hand, the distribution of magnitude for the case III is more diffuse and thus a loss function with large weighting was required for finding the optimum circuit parameters.

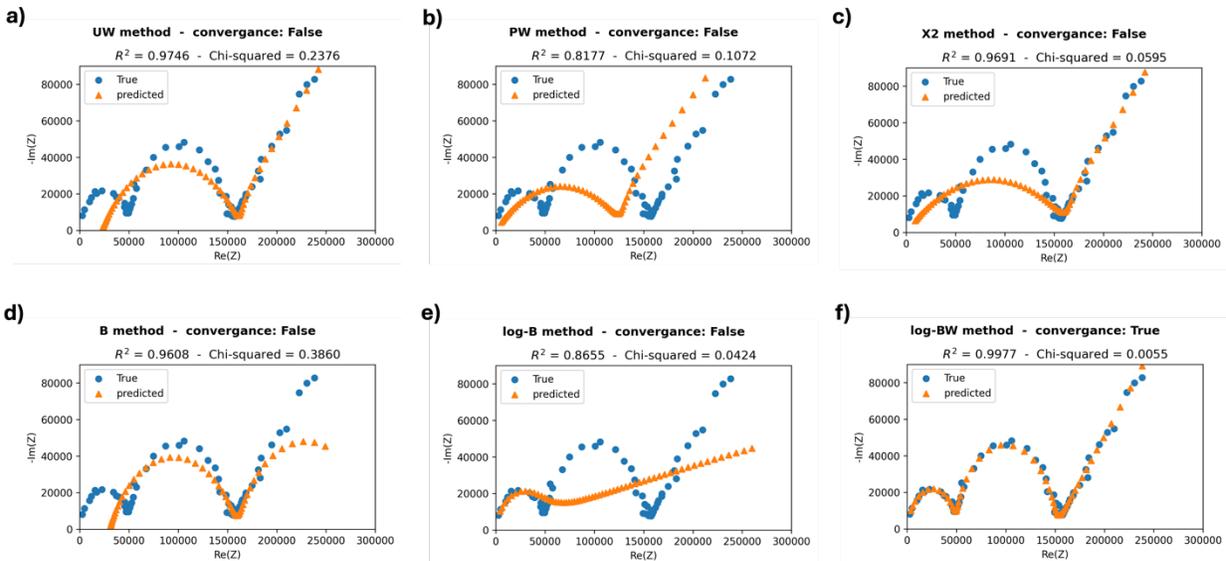

Figure 4: The true and the ECM-predicted impedance data fitted with different loss functions for circuit of Figure 1-c: a) UW, b) PW, c) X2, d) B, e) log-B, and f) log-BW.

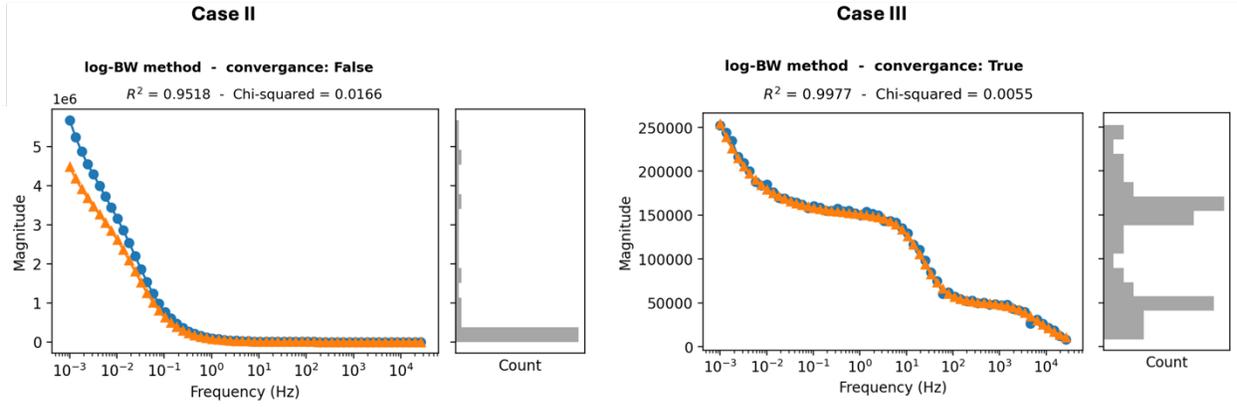

Figure 5: The magnitude Bode plot of case II and case III with the distribution of magnitudes.

### 3.2. Statistical comparison for each loss function

As discussed in the previous section, there may exist some cases where the convergence threshold was never met before reaching the maximum allowable iteration. Figure 6 shows the number of EIS data that met the threshold for each loss functions used in NLS fitting. Both unweighted loss functions (i.e. UW and B) showed the smallest number of converged cases while the X2 and log-B ranked the first and second, respectively. For the UW and B loss functions, only 37% and 33% of the datasets met the convergence threshold, respectively. In contrast, 98.7%, 98.2%, 97.9%, and 96.5% of the datasets satisfied the threshold when X2, log-B, PW, and log-BW, were used respectively. The difference in convergence can be attributed to the fact that the choice of loss function influences the convexity of the loss landscape, thereby affecting the optimization process and convergence. [22] Therefore, for the rest of the paper we focused on only the weighted loss functions (the first four blue bars in Figure 6) and compare the results on the EIS data that mutually achieved the convergence among all weighted loss functions.

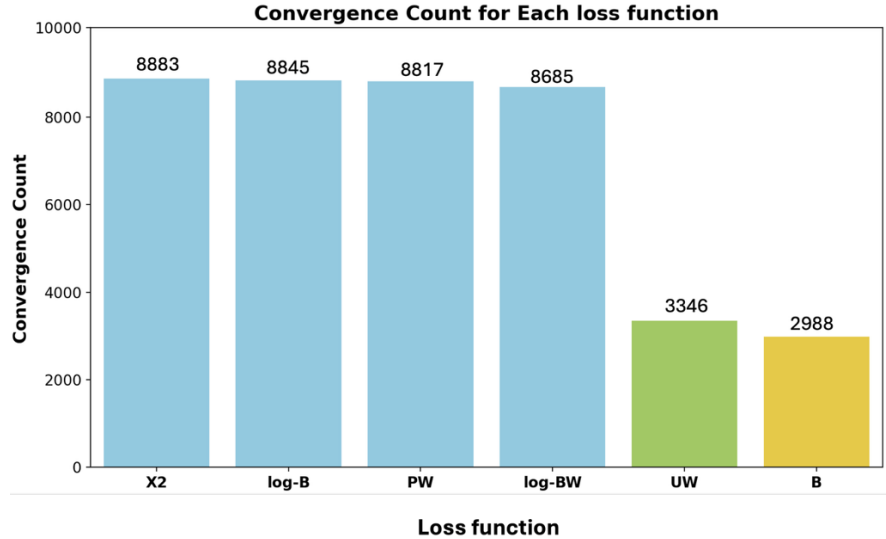

Figure 6: Number of EIS data that met the convergence threshold during the least square fitting process. The first four blue bars from the left belong to the loss functions considered in the rest of the paper.

### 3.2.1. Statistical comparison of performance metrics

The box plots of performance metrics and the distribution of chi-squared are illustrated in Figure 7 for EIS data fitted with different loss functions. In these plots, the red circles indicate the mean values of each metric for the corresponding loss function. These plots clearly show that the X2 and log-B loss functions outperformed the other two with a slightly better performance for X2. This suggests that the log transformation in log-B could correct the bias related to the scale during the least square procedure even though no weight was explicitly included in this loss function formulation.

In Figure 7, the X2 box plots for $R^2$_score and $R^2$_magnitude metrics are narrower and more skewed toward higher values. However, in terms of $R^2$_phase, the log-B yielded a slightly narrower distribution. In addition, X2 loss function yielded marginally lower chi-squared values. The mean of the chi-squared for the data fitted by X2 is 0.00043 while for log-B the mean of chi-squared slightly increased by 2% to 0.00044. Relative to this slight improvement in the fit quality, the computational cost of X2 is quite higher than that of log-B. The mean computational time for log-B is around 1.05 seconds while it is 1.45 seconds for X2 (i.e., 1.4 times slower for X2). In addition, the time box plot for log-B is remarkably narrower than the other loss functions. Thus, the X2 loss function may be more suitable when the quality of fit is a priority. However, when computational efficiency is a concern, the log-B loss function is recommended, as it provides a comparable quality of fit while reducing computational cost.

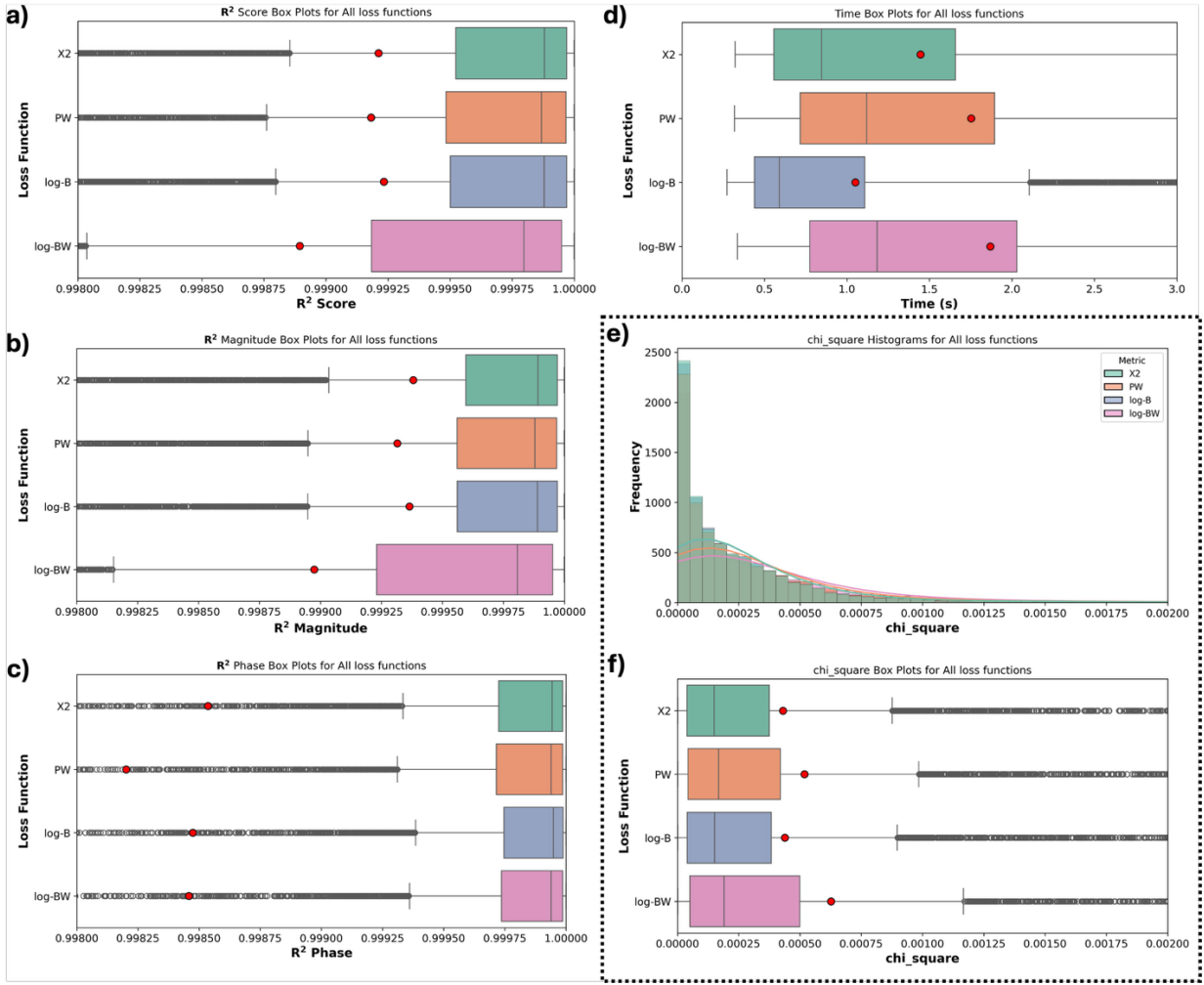

Figure 7: Box plots of a) $R^2$_score, b) $R^2$_magnitude, c) $R^2$_phase, d) time for EIS data fitted with different loss functions in least square fitting. The distribution and box plot for chi-square are shown in e) and f), respectively. The red circles show the average values. In these plots the x-axis was truncated for better visualization purpose and so some outliers may not be visible.

### 3.2.2. Difference between the predicted and original component values

To compare the predicted and original component values, a pre-screening step was performed on the dataset following the fitting procedure to ensure that the predicted parameters converged to the same local minimum as the ground truth. The rationale behind this step is that multiple sets of component values can produce equally good fits and meet the fitting threshold while differing significantly from the original values. In this case, the difference between the predicted and original component values becomes significant even though both solutions provide very high quality of fit. To address this, the data where the absolute percentage error (APE) of any component parameters exceeds 100% was excluded from the analysis. Figure 8 shows the percentage of the data retained after this pre-screening step. As shown in the figure, for all loss functions except log-BW, more than 95% of the data remained after pre-screening for most circuits, while R1-[P2,R3]-P4 and R1-[P2,R3]-[P4,R5]-P6 retained only about 70% and 40%, respectively. In contrast, log-BW retained substantially fewer cases across most circuits, indicating that this loss function often converges to solutions deviating from the original component values.

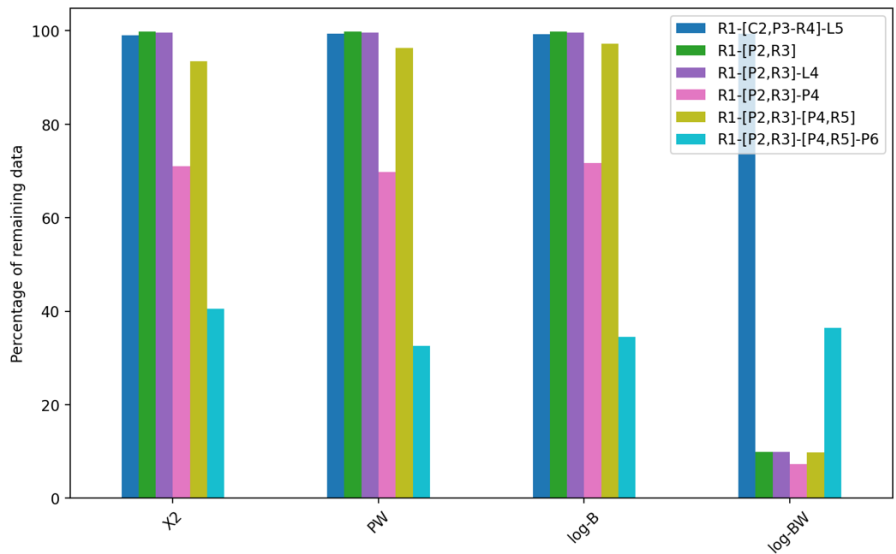

Figure 8: Percentage of the remining EIS data fitted with different loss functions after the pre-screening step to exclude the data where the APE for any component of the circuit exceeds 100%.

The MAPE for each component in the circuit was calculated using the screened data and is illustrated in Figure 9. In this figure, *Pw* and *Pn* denote the magnitude and $\alpha$ of the constant phase element, respectively. This figure clearly shows that the MAPE is below 3% for all the components in all circuits except for circuit 'd', where some components exhibit MAPE values approaching 10%. For circuits 'a' and 'c', the log-B loss function consistently achieved the lowest MAPE across all the components. In circuit 'b', eighter log-B or log-BW resulted in the lowest MAPE across all components. For the remaining circuits, the best-performing loss function varied by the component type. Therefore, for an overall comparison, the average of MAPE across all the components was calculated for each loss function and is illustrated on the right side of each plot in Figure 9 with the label "Average." Among all the loss functions, log-B achieved the lowest average MAPE across all circuits, except in circuit 'e', where log-BW performed best. This highlights the effectiveness of log-B for finding component values closer to the original parameters used during data generation.

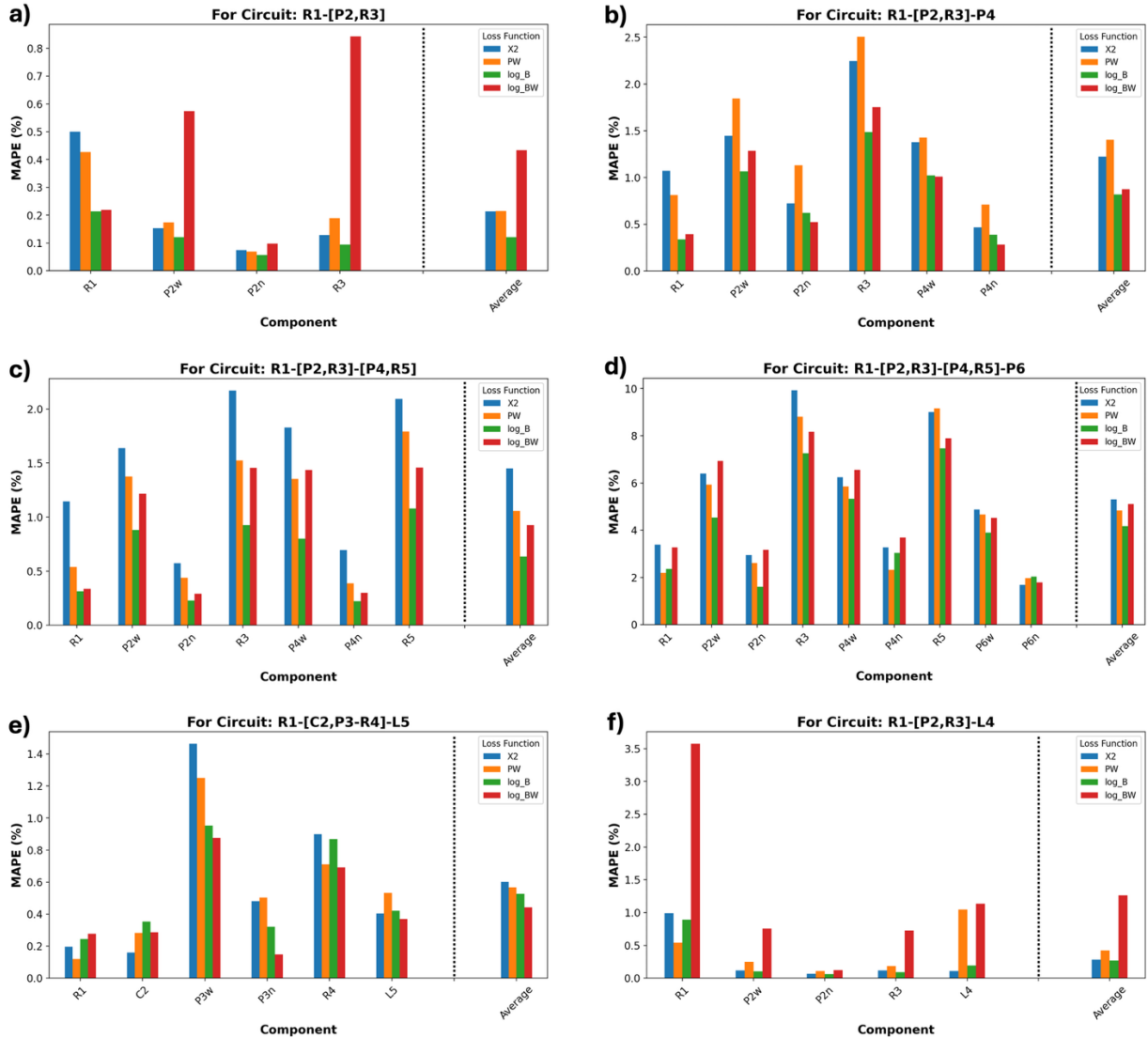

Figure 9: Mean Absolute Percentage Error (MAPE) of each component parameter across different circuits, fitted using various loss functions. *Pw* and *Pn* represent the magnitude and α of the constant phase element, respectively. The final label on the right side of each plot, titled "Average", indicates the average MAPE across all component parameters for each loss function.

### 3.2.3. Choosing the appropriate Loss Function

By comparing all the results, we can now provide insights into the appropriate selection of loss functions under different conditions. Overall, if the maximum allowable iteration is limited in NLS fitting algorithm then one should choose X2 as Figure 6 showed the largest number of convergences for this loss function. The percentages of the EIS data that satisfied the convergence threshold are 98.7, 98.2, 97.9, and 96.5% for X2, log-B, PW, and log-BW, respectively. If the computational cost matters the most, we recommend choosing the log-B loss function, as it provides comparable quality of fit with 1.4 times faster NLS fitting than X2. Although this computational cost reduction may seem minor, it becomes crucial when performing large-scale NLS fitting for data-driven approaches, such as training machine learning models on extensive datasets or for large iterations.

If the quality of fit is the primary concern, then the choice of loss function depends on the performance metrics as shown in **Table 2**. In this table, we summarized the mean of each performance metric and highlighted the cell that yielded the best performance for each respective metric. Based on the mean values, we recommend using X2 for achieving the highest quality of fit in terms of chi-squared, $R^2$_phase, and $R^2$_magnitude. However, in terms of $R^2$_score, one should choose log-B. We should emphasize that, although the comparison is made at the fourth or fifth decimal place, such differences can still be significant when evaluating the performance of machine learning models.

**Table 2:** The mean of performance metrics for each loss functions used to fit ECMs for both wide and narrow EIS data. The highlighted cells show belong to the loss functions that outperforms the others for each performance metrics

| Performance metrics | Loss functions | | | |
| --- | --- | --- | --- | --- |
| | X2 | PW | log-B | log-BW |
| chi-squared | 0.00043 | 0.00051 | 0.00044 | 0.00063 |
| R2_score | 0.99921 | 0.99918 | 0.99923 | 0.99889 |
| R2_magnitude | 0.99938 | 0.99931 | 0.99936 | 0.99897 |
| R2_phase | 0.99853 | 0.99820 | 0.99847 | 0.99845 |
| time | 1.4458 | 1.7527 | 1.0504 | 1.8684 |

While the table above shows the winner loss function for each performance metric, the relative improvements of one metric over the other by changing the loss function could be a key selection criterion. So, we normalized each performance metric for all loss functions and plotted a radar chart in Figure 10. Since lower values of time and chi-squared indicate better performance (in contrast to $R^2$ where higher values are better), we applied inverse normalization to these two metrics to ensure consistent interpretation across the radar chart. In this figure, the closer the edges to the unity, the higher the overall performance of the loss function (i.e., larger area).

In Figure 10, the main two competitors in terms of the size of the area are X2 and log-B. In contrast, the areas of log-BW and PW are significantly smaller and cannot be the candidate loss function. In terms of fit quality, X2 outperformed log-B by 10% in $R^2$_magnitude and chi-squared, and 20% in $R^2$_phase, while log-B performed 10% better in $R^2$_score. Even though the quality of fit is relatively close, the computational efficiency of log-B is almost 50% larger than X2, causes a significant increase in the area associated to this loss function in the radar chart. Therefore, log-B becomes a very strong candidate relative to other loss functions.

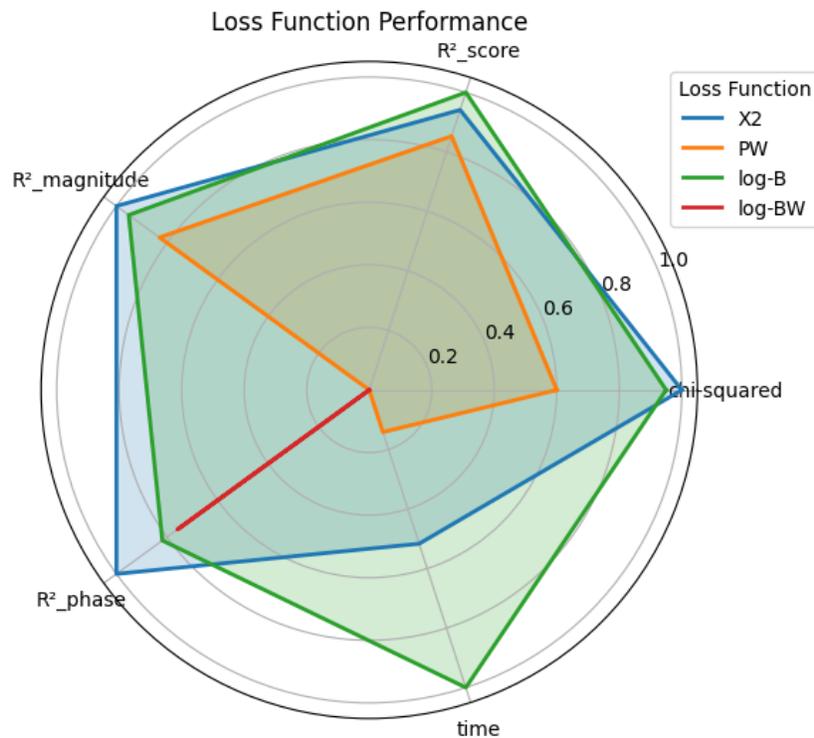

Figure 10: Radar plots of normalized performance metrics for different loss functions.

# 4. Conclusions

The performance of different loss functions for fitting the ECM parameters to the EIS data was investigated. The initial analysis focused on three case studies to examine the effects of magnitude scale range and data distribution. We studied the cases where the loss functions with strong weighting did not yield better performance and showed that both range and distribution of impedance magnitudes may influence the selection of an appropriate loss function. We also showed that different performance metrics are necessary for the assessment of fit quality. Subsequently, we conducted a more statistical analysis using a large set of synthetic EIS data. The X2 loss function demonstrated the highest convergence rate and superior fit quality across several performance metrics. This analysis revealed that X2 yielded 5-20% higher performance across all metrics, except 5% lower R2_score. However, this improvement comes at a higher computational cost, which becomes especially important in data-driven applications. In contrast, log-B provided comparable quality of fit while being 50% more computationally efficient and yielding lower MAPE for component values in most circuits. This balance between accuracy and efficiency positions log-B as a strong alternative, particularly in data-driven applications such as training machine learning models on extensive datasets or iterations. These insights provide practical guidance for selecting loss functions in fitting the EIS data to ECM and can facilitate more reliable and scalable solutions for analysis of electrochemical systems.

# 5. Acknowledgments

This research was funded by the Government of Canada under the National Research Council Canada's Collaborative Science, Technology and Innovation Program (CSTIP)'s Critical Battery Materials Initiative (CBMI).

## Supplementary Materials

In addition to least-squares fitting as the local optimization method, we also evaluated basin-hopping as a global optimization approach to assess the performance of different loss functions. Figure S.1 shows the number of EIS data that met the convergence threshold for each loss function used in basin-hopping fitting. Only about 20% of the data converged, compared to 96–98% with least-squares fitting as shown in Figure 6. This low convergence rate highlights the instability of basin-hopping, making it impractical for data-driven applications where both reliability and efficiency are critical.

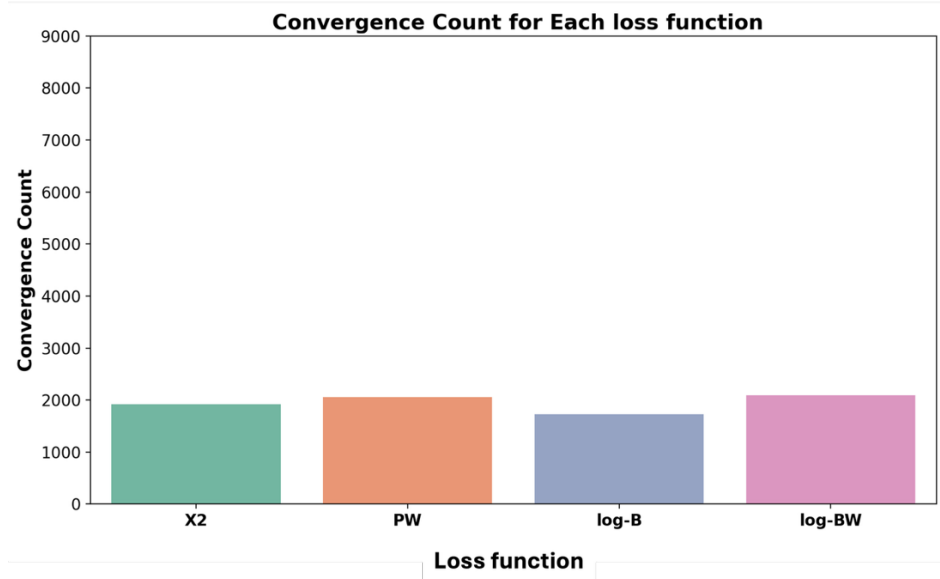

Figure S.1: Number of EIS data that met the convergence threshold during the basin-hopping fitting process.

We also performed the same statistical analysis as in Section 3.2 for the loss functions used in basin-hopping fitting. The results are shown in Figure S.2. Compared to Figure 7, the box plots are much wider, particularly for $R^2$_score and $R^2$_magnitude, indicating greater variability and lower quality of fit than when using least-squares fitting. In addition, the computational cost of basin-hopping is about 40 times higher: while least-squares fitting averages around 1.5 seconds across all loss functions, basin-hopping requires around 60 seconds. This substantial computational cost makes basin-hopping impractical for large-scale fitting in data-driven applications, such as training machine learning models on extensive datasets or performing many iterations. These results show the inefficiency and instability of basin-hopping for data-driven applications, which is why this study focused on least-squares fitting as the local optimization method.

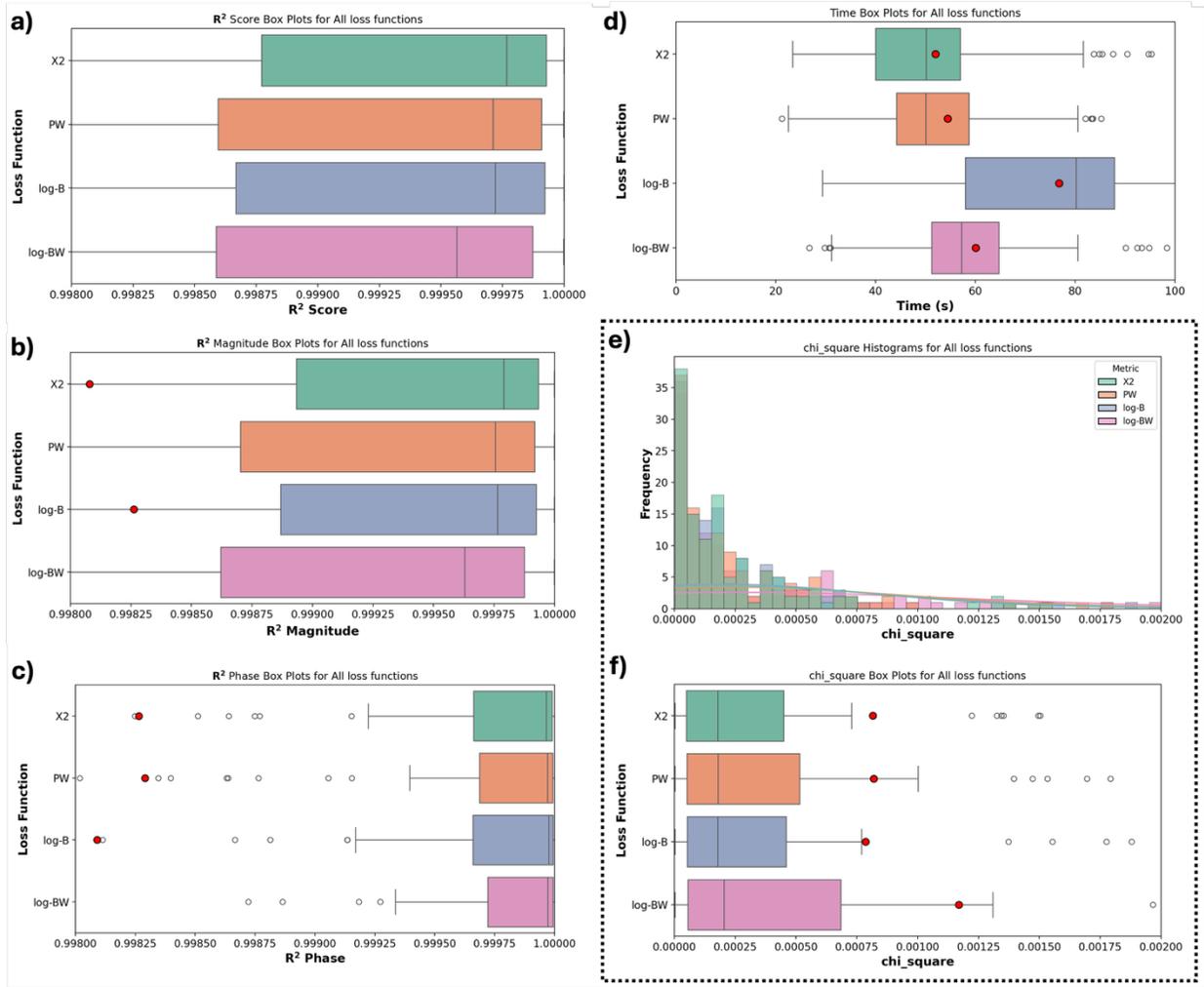

Figure S.2: Box plots of a) R2_score, b) R2_magnitude, c) R2_phase, d) time for EIS data fitted with different loss functions in basin-hopping fitting. The distribution and box plot for chi-square are shown in e) and f), respectively. The red circles show the average values. In these plots the x-axis was truncated for better visualization purpose and so some outliers may not be visible.